\documentclass[lettersize,journal]{IEEEtran}
\usepackage{amsmath,amsfonts}
\usepackage{algorithmic}
\usepackage{algorithm}
\usepackage{array}
\usepackage[caption=false,font=normalsize,labelfont=sf,textfont=sf]{subfig}
\usepackage{textcomp}
\usepackage{stfloats}
\usepackage{url}
\usepackage{verbatim}
\usepackage{graphicx}
\usepackage{cite}
\usepackage{makecell}
\usepackage{graphicx}
\usepackage{booktabs}
\usepackage{multirow}
\usepackage[table]{xcolor}
\usepackage{pifont}
\newcommand{\cmark}{\ding{51}}%
\newcommand{\xmark}{\ding{55}}%
\definecolor{Gray}{gray}{0.8}
\definecolor{LG}{gray}{.83}
\definecolor{het}{RGB}{255,245,235}
\usepackage[dvipsnames]{xcolor} 
\colorlet{PastelSky}{SkyBlue!70!white}      
\colorlet{PastelDeepSky}{SkyBlue!60!blue!75!white} 
\colorlet{PastelSkyLighter}{SkyBlue!10!white} 
\usepackage{caption}
\usepackage{xspace}
\newcommand{\eg}{\textit{e.g.}\xspace}

\begin{document}

\title{HEAT: Heterogeneous End-to-End Autonomous Driving via Trajectory-Guided World Models}

\author{
Hoonhee Cho$^{*1}$,
Giwon Lee$^{*1}$,
Jae-Young Kang$^{*1}$,
Hyemin Yang$^{*1}$,
Heejun Park$^{*1}$,
and Kuk-Jin Yoon$^{\dagger 1}$%
\thanks{
$^{*}$Equal contribution. $^{\dagger}$Corresponding author.
}%
\thanks{
$^{1}$All authors are with KAIST, Daejeon, Republic of Korea.
}%
}

\markboth{Journal of \LaTeX\ Class Files,~Vol.~14, No.~8, August~2021}%
{Shell \MakeLowercase{\textit{et al.}}: A Sample Article Using IEEEtran.cls for IEEE Journals}


\maketitle

\begin{abstract}
End-to-end autonomous driving has emerged as a compelling alternative to traditional modular pipelines by directly mapping raw sensor data to driving actions. While recent approaches achieve strong performance on single-domain datasets, their performance degrades significantly when trained jointly across multiple heterogeneous domains. In practice, however, autonomous systems must operate across diverse environments with heterogeneous distributions, including different cities, sensor configurations, and traffic patterns, without domain-specific retraining.
This gap highlights a key challenge in multi-domain learning: domain-specific variations across heterogeneous domains introduce conflicting learning signals, driving models toward compromised solutions that are suboptimal across domains. To address this, we propose a trajectory-driven learning paradigm that organizes training around planning trajectories, enabling the model to capture domain-invariant representations of driving intent. Furthermore, we incorporate a world model that predicts future latent features conditioned on ego actions, improving feature consistency and mitigating domain-induced biases.
We evaluate our approach on three benchmarks, nuScenes, NAVSIM, and the Waymo end-to-end dataset, and show substantial improvements over existing methods across all domains. Our results demonstrate that a single unified model can be trained on heterogeneous datasets while maintaining strong performance within each domain, highlighting a step toward scalable real-world deployment. We will make our code publicly available.



\end{abstract}

\section{Introduction}
\label{sec:intro}

\begin{figure*}[t]
    \centering
    \includegraphics[width=0.99\textwidth]{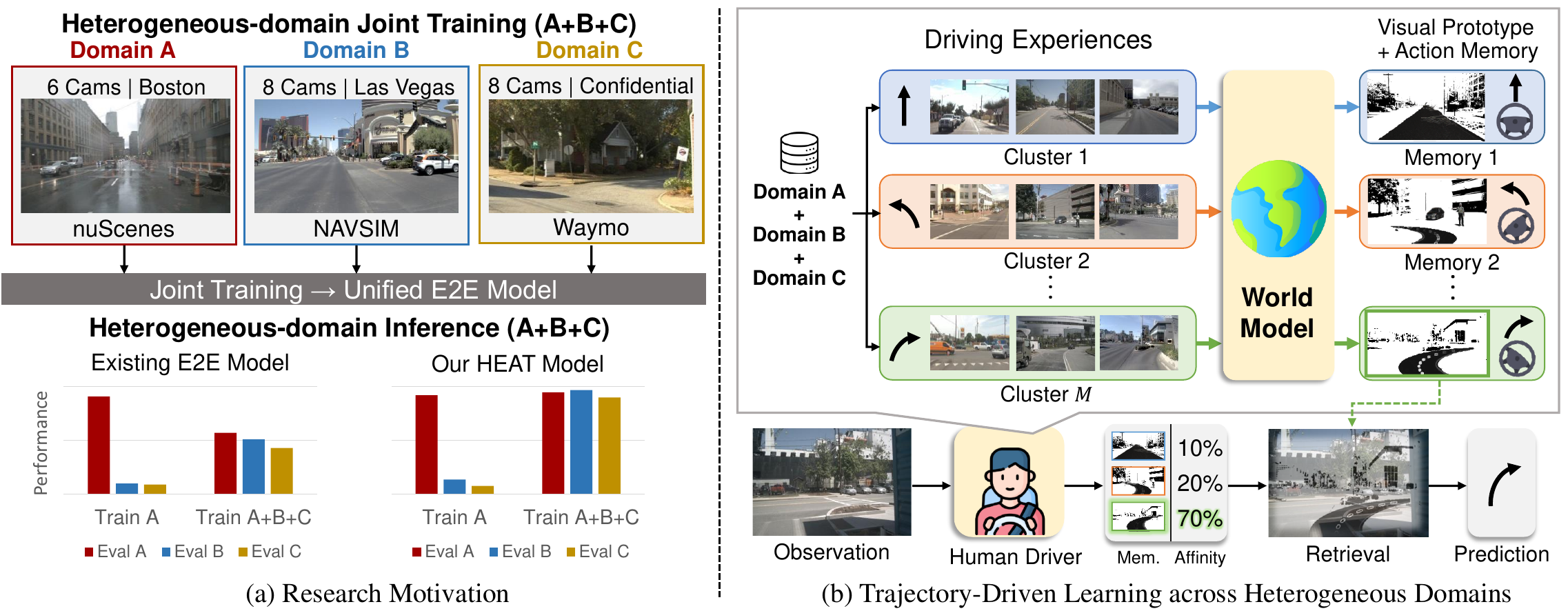}
    \vspace{-5pt}
    \caption{In real-world deployments, E2E-AD inevitably encounters heterogeneous domains with diverse data distributions. Our objective is to develop a model that generalizes robustly across such variability. 
    Motivated by the way human drivers recall and adapt prior experiences to new situations, we introduce a trajectory-driven world model designed to transfer behavioral knowledge across domains.
    }
    \label{fig:teaser}
    \vspace{-11pt}
\end{figure*}


\IEEEPARstart{E}{nd-to-end} autonomous driving (E2E-AD)~\cite{E2E_survey_IEEE2024} is a strong alternative to modular pipelines. In contrast to conventional planners~\cite{note2eplan_8_CVPR2025_zhang2025carplanner} that rely on pre-processed perception outputs (\eg,~3D bounding boxes), E2E-AD~\cite{E2E_imitation_learning_ICRA2018} uses raw sensor data to learn scene-level features. This reduces information loss, simplifies the pipeline, and improves robustness by tightly coupling sensor information and decision-making.
Although promising empirical results have been reported, the effectiveness of E2E-AD under heterogeneous multi-domain settings remains underexplored, as most studies still evaluate on single-domain datasets. In practice, autonomous vehicles must scale without bespoke retraining for every car model or city, requiring models to operate across heterogeneous distributions spanning different road layouts, hardware (\eg,~camera setups~\cite{E2E_21_gaussian_NeurIPS2025_cho2025vr}), and traffic patterns.
However, as shown on the left in Fig.~\ref{fig:teaser}, simply training on a mixture of datasets from diverse distributions already poses significant challenges, as current end-to-end models often exhibit performance degradation when exposed to heterogeneous data. This highlights fundamental limitations in scalability and robustness under multi-domain training settings. In this context, generalization to entirely unseen domains remains out of focus, as achieving stable performance across known heterogeneous domains is itself a non-trivial problem.
Motivated by this gap, we investigate E2E-AD under multi-domain training settings, where data originate from diverse distributions, and ask whether a single model can be trained jointly across multiple domains while maintaining strong performance within each domain.

However, such a formulation often drives the model to rely on domain-specific signals, leading to compromised representations that are suboptimal across domains.
To this end, we turn our attention to the underlying structure of driving behaviors rather than domain-specific appearances. Instead of treating each dataset as fundamentally different due to variations in city layouts, camera setups, or visual contexts, we cluster data according to their planning trajectories. This perspective is motivated by a simple intuition: visually different scenes may correspond to the same driving situation when the required behavior (\eg, turning left or go straight) is similar. By learning from planning trajectories, E2E-AD model can capture domain-invariant representations of driving intent, focusing on the underlying behavior rather than superficial visual differences.


As illustrated on the right in Fig.~\ref{fig:teaser}, we further strengthen this process by employing a world model~\cite{min2024driveworld,E2E_LAW_ICLR2025}, inspired by the human ability to imagine and reason about possible futures while driving.
We find that the world model not only improves the quality of the visual feature space and facilitates multi-domain fusion, but also conditions on actions, tightly coupling visual representations with action signals to enable more coherent learning of both visual representations control policies.
By combining the proposed planning-based learning paradigm with the use of a world model, the proposed novel heterogeneous framework, named HEAT, can treat datasets with different distributions as a single unified dataset. This synergy allows the model to focus on the shared behavioral structure of driving rather than domain-specific variations, effectively narrowing the gap between heterogeneous sources.

To validate this idea, we conduct extensive experiments on three representative real-world benchmarks: nuScenes~\cite{Benchmarks_NuScenes}, NAVSIM~\cite{dauner2024navsim} (built on nuPlan~\cite{caesar2021nuplan}), and the Waymo end-to-end dataset~\cite{waymo2025e2e}. We jointly employ them as a heterogeneous training and evaluation suite, reflecting the variety of conditions encountered in real-world driving. The results highlight the potential of designing a unified E2E-AD model that performs robustly across diverse domains and paves the way toward practical deployment. 
Our main contributions are summarized as follows:
\begin{itemize}
\item
\textbf{Trajectory-driven representation learning.} We propose a novel learning paradigm that organizes training around planning trajectories. By clustering data based on behavioral similarity and constructing prototypes and action memory, our framework encourages representations that capture driving behavior beyond domain-specific appearances.

\item
\textbf{Heterogeneous-domain benchmark with superior performance.}
We establish a standardized benchmark setting for an end-to-end autonomous driving under heterogeneous domains, enabling fair and reproducible evaluation across diverse datasets. Under this benchmark, our framework achieves strong multi-domain generalization and robustness, substantially surpassing other methods and setting a solid reference point for future research.

\end{itemize}

\section{Related Works}
\label{sec:related_works}

\noindent
\textbf{End-to-End Autonomous Driving (E2E-AD).}
E2E-AD maps raw sensor inputs directly to driving decisions.
Early works improved planning by optimizing submodules~\cite{E2E_UniAD,E2E_Transfuser++} and learning spatiotemporal features~\cite{E2E_ST-P3}.
Parallel planners enhance efficiency~\cite{E2E_paradrive}, while structured planning vectorizes lanes and agents~\cite{E2E_VAD,E2E_vadv2}.
A complementary direction~\cite{E2E_LAW_ICLR2025,zheng2025world4drive,  E2E_ADMLP_archeive,E2E_BEVPlanner,E2E_Tranfuser, E2E_TCP} removes explicit perception supervision by learning latent or self-supervised representations.
Despite recent progress in E2E-AD, most methods are evaluated on single domains, leaving robustness across heterogeneous environments underexplored. In practice, autonomous systems must operate across diverse environments without retraining. However, learning a unified model for heterogeneous data is challenging, as domain variations introduce conflicting signals that degrade joint training. We therefore study multi-domain E2E-AD and examine whether a single model can maintain strong performance across heterogeneous distributions.

\noindent
\textbf{World Model in Autonomous Driving.}
World models predict future scene states under candidate controls. Image-centric variants~\cite{WORLD_DriveWM} learn latent visual dynamics and generate controllable long-horizon videos. Occupancy-based methods model 3D world states for forecasting and planning~\cite{E2E_OccNet_ICCV2023}. Recent work integrates prediction with planning, including BEV forecasting for trajectory selection~\cite{E2E_WORLD_PLAN_li2025end_wote} and navigation-guided tokens for efficient control~\cite{E2E_E2EAD_ICLR2025_navguided}. 
Latent world models instead predict future latents for downstream querying without heavy geometric supervision~\cite{E2E_LAW_ICLR2025,zheng2025world4drive}. 
Despite these advances, most work remains single-domain, leaving multi-domain planning underexplored. We address this via world modeling for heterogeneous-domain generalization.

\section{Methodology}
\label{sec:method}

\subsection{Task Formulation}

In vision-based E2E-AD, the objective is to plan the ego-vehicle’s future trajectory as a sequence of waypoints. Let $\mathcal{I}_t=\{I_t^{1}, I_t^{2}, \ldots, I_t^{K}\}$ denote $K$ synchronized surround-view images captured at time $t$. Conditioned on $\mathcal{I}_t$, the model predicts $T$ waypoints in the BEV plane. Specifically, the output is $\mathbf{W}_t=\{\mathbf{w}^{\,1}_t,\mathbf{w}^{\,2}_t,\ldots,\mathbf{w}^{\,T}_t\}$, where each waypoint $\mathbf{w}^{\,i}_t=(x^{\,i}_t,y^{\,i}_t, \phi^{\,i}_t)  \in \mathbb{R}^3$ represents the predicted BEV position and heading angle of the ego vehicle at time $t+i$. In compact form, we model this mapping as $\mathbf{W}_t = f_\theta(\mathcal{I}_t)$.
Following recent advances~\cite{E2E_LAW_ICLR2025, zheng2025world4drive}, we adopt a perception-free formulation that directly maps raw sensor inputs to driving actions without relying on explicit perception supervision. This design avoids dependence on costly annotations and provides greater flexibility when learning from heterogeneous data, making it well-suited for multi-domain training settings.
The encoder produces a set of visual feature tokens $\mathbf{V}_t = \{\mathbf{v}_t^1, \mathbf{v}_t^2, \ldots, \mathbf{v}_t^L\}$, where each token $\mathbf{v}_t^l \in \mathbb{R}^D$ encodes local scene information and $L$ denotes the number of feature tokens. These visual tokens are then processed by a transformer module to predict the future trajectory $\mathbf{W}_t$.

To account for domain heterogeneity, we consider multiple datasets
$\mathcal{D}=\{\mathcal{D}_1,\mathcal{D}_2,\ldots,\mathcal{D}_N\}$ drawn from different distributions.
Our objective is to learn a single mapping function $f_\theta$ that generalizes across heterogeneous domains and achieves consistently high performance.
Formally, given a sample $\mathcal{I}_t^{(d)}\! \sim \! \mathcal{D}_d$ from domain $d \in \{1,\ldots,N\}$, the E2E-AD model predicts waypoints
$\mathbf{W}_t^{(d)} = f_\theta(\mathcal{I}_t^{(d)})$.
In our setup, we specifically consider $N=3$ domains: nuScenes, NAVSIM, and Waymo, each representing a distinct distributional condition, and train the model jointly over all datasets.

\subsection{Overall Framework}

\begin{figure*}[t]
    \centering
    \includegraphics[width=.99\textwidth]{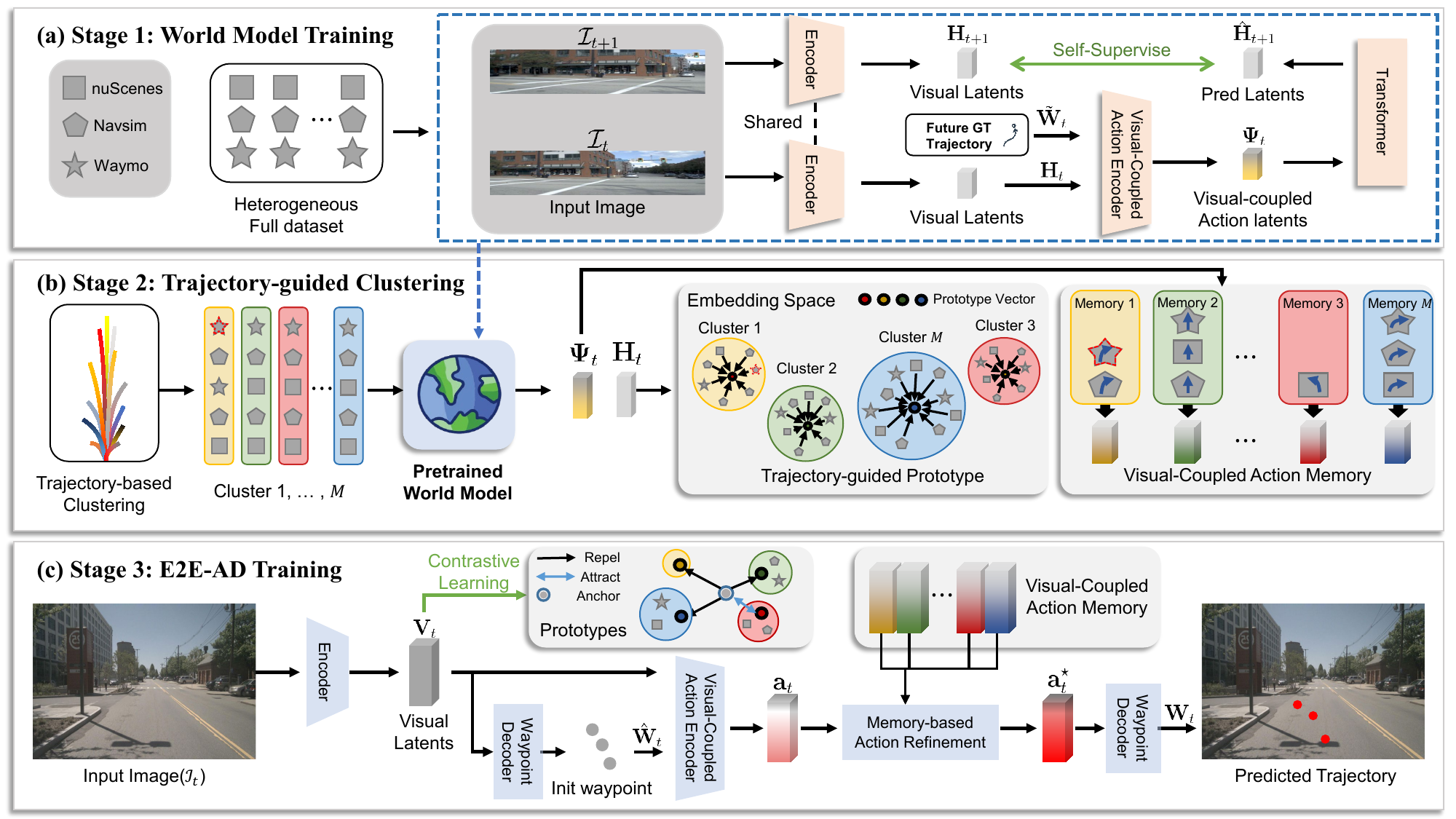}
    \vspace{-6pt}
    \caption{Overview of HEAT. We first pretrains a trajectory-conditioned world model to learn trajectory-aligned representations that couple visual observations with future actions. The resulting latents are organized into domain-invariant priors via trajectory-guided behavior clustering, producing visual prototypes and an action memory. Finally, an E2E-AD model is trained using these priors through contrastive alignment and episodic memory refinement, enabling robust multi-domain driving.}
    \label{fig:overall_method}
    \vspace{-6pt}
\end{figure*}

The proposed framework, HEAT, mitigates domain bias and reduces overfitting to specific environments through trajectory-driven representation learning. Inspired by how humans link observations to potential future actions, HEAT leverages a world model to learn a generalized representation where visual observations and actions are tightly coupled across heterogeneous domains. As illustrated in Fig.~\ref{fig:overall_method}, the framework consists of three stages.
In the first stage (Sec.~\ref{sec:world_model}), we pretrain a trajectory-conditioned world model to learn a trajectory-aligned latent space where visual observations are coupled with future actions. This stage focuses on learning generalizable representations via trajectory-driven supervision, and the world model is used only during pretraining.
In the second stage (Sec.~\ref{sec:cluster}), we extract latents from the pretrained world model and organize them into structured domain-invariant priors. We perform trajectory-guided behavior clustering using ground-truth waypoints and summarize each cluster into (i) trajectory-guided visual prototypes and (ii) a visual-coupled action memory. These structures guide the transfer of trajectory-relevant knowledge across heterogeneous domains without relying on appearance cues.
In the third stage (Sec.~\ref{sec:e2e_model}), we train E2E-AD model from scratch using the structured priors from the second stage. The model is aligned with trajectory-guided prototypes via a contrastive objective and further refined with episodic action memory to encourage action-consistent predictions, enabling effective multi-domain learning.

\subsection{Stage1: World Model for Trajectory-Driven Learning}
\label{sec:world_model}
To mitigate domain bias, in which models overfit to environment-specific cues (\eg,~visual appearance), we learn representations through the more general concept of trajectories. We adopt a world model~\cite{E2E_WORLD_PLAN_hassan2025gem, min2024driveworld, WORLD_PLAN_ECCV2025_nemo, E2E_WORLD_PLAN_chen2024drivinggpt} that captures the dynamics linking observations and future actions. Our world model conditions on ground-truth trajectories and infers future sensor inputs in the visual latent space, which tightly couples the visual and action latents and directly supports trajectory-driven learning. In our framework, the world model is used for pre-training to obtain trajectory-coupled, domain-agnostic representations that are subsequently transferred to the E2E-AD model, thereby enhancing action-relevant feature extraction without reliance on domain-specific information.

\noindent
\textbf{World Model Training.}
To align the representation space with future trajectories, we design a latent world model~\cite{E2E_LAW_ICLR2025} that integrates current visual features with ground-truth waypoints. Given the surround-view images \(\mathcal{I}_t\), the encoder produces a set of visual latent feature tokens 
\(\mathbf{H}_t = \{\mathbf{h}_t^1, \mathbf{h}_t^2, \ldots, \mathbf{h}_t^L\}\), where each token \(\mathbf{h}_t^l \in \mathbb{R}^D\) encodes local scene information and $L$ denotes the number of latent tokens. For conditioning, the future ground-truth trajectory 
\(\mathbf{\tilde{W}}_t = \{\mathbf{\tilde{w}}_t^1, \mathbf{\tilde{w}}_t^2, \ldots, \mathbf{\tilde{w}}_t^T\}\) 
is flattened into a vector \(\mathbf{\tilde{w}}'_t \in \mathbb{R}^{3T}\). 
This vector is concatenated with each latent token \(\mathbf{h}_t^l\) to construct 
visual-coupled action latents:  
\begin{equation}
\psi_t^l = \text{MLP}\big([\mathbf{h}_t^l, \mathbf{\tilde{w}}'_t]\big), 
\quad l=1,\ldots,L,
\end{equation}
where \([\cdot,\cdot]\) denotes channel-wise concatenation. The set of visual-coupled action latents \(\Psi_t =\{\psi_t^l\}_{l=1}^L\) is then processed by transformer blocks~\cite{vaswani2017attention} to predict the representation of the next frame:  
\begin{equation}
\hat{\mathbf{H}}_{t+1} = \text{Transformer}(\Psi_t).
\end{equation}
During training, the world model is self-supervised using visual latents of the subsequent frame extracted from images at time $t+1$. To prevent representation collapse, we follow~\cite{E2E_LAW_ICLR2025} and impose an additional supervision to predict future trajectories from visual latents.
By minimizing the discrepancy between predicted and ground-truth latents with a mean squared error objective, the model learns a richer and more structured representation that captures future trajectories with higher fidelity. Moreover, instead of simply predicting future visual latents, it uses trajectory-guided learning to better align visual and action latents, resulting in a tighter perception–control coupling.

\subsection{Stage2: Trajectory-guided Behavior Clustering}
\label{sec:cluster}

We aim to extract well-represented features from a pre-trained world model and use them to construct domain-invariant representations, explicitly mitigating domain bias while promoting trajectory-based learning.
Specifically, we define the behavior set by gathering the behavior triplets from all samples in the union of the training sets $\mathcal{D}$, resulting:
$\mathbf{B} = \{ (\mathbf{H}_t^{(d)}, \mathbf{\Psi}_t^{(d)}, \mathbf{\tilde{W}}_t^{(d)} ) \} $.
We apply K-Means to the behavior set $\mathbf{B}$ based on their ground-truth trajectory waypoints $\tilde{\mathbf{W}}_{t}$, grouping them into $M$ behavior-centric clusters:
\begin{equation}
\scalebox{0.9}{$
\mathcal{C}_m = \Big\{ \big(\mathbf{H}_t, \mathbf{\Psi}_t, \mathbf{\tilde{W}}_t \big) 
\;\Big|\; \mathbf{\tilde{W}}_t \in \text{cluster } m \Big\}, ~m=1,\ldots,M
$}
\end{equation}

\noindent
\textbf{Trajectory-guided Prototype from World Model.}
To achieve domain-invariant learning across heterogeneous datasets, 
it is necessary to abstract away domain-specific variations 
while retaining trajectory-relevant semantics. 
A natural solution is to summarize trajectory-aware features into representative prototypes, 
which serve as compact and transferable knowledge across domains. 
After forming the clusters, we aggregate the visual latents into such prototypes. For each cluster \(\mathcal{C}_m\), the corresponding 
visual latent $\mathbf{H}_t  \in \mathbb{R}^{L \times D}$
from all samples $s$ in the cluster are averaged to compute a prototype:  

\begin{equation}
\mathbf{P}_m \;=\; \frac{1}{\left|\mathcal{C}_m\right|}\sum_{s \in \mathcal{C}_m} \mathbf{H}_t^{(s)} ,
\end{equation}
where the resulting prototype \(\mathbf{P}_m \in \mathbb{R}^{L \times D}\) preserves the latent token 
dimension \(L\). This ensures that each cluster is represented not by a single vector, 
but by a rich token-level prototype set that captures trajectory-aligned structure 
while suppressing domain biases. These prototypes later guide the transfer to the 
end-to-end autonomous driving model.

\noindent
\textbf{Visual-Coupled Action Memory from World Model.}
Inspired by how humans ground present decisions in past, visually similar experiences, we store action latents in a dedicated memory keyed by visual context, enabling efficient retrieval of action features for the current driving scenario.

Specifically, for each cluster $\mathcal{C}_m$, where $m \in \{1,\dots,M\}$ and $M$ denotes the number of behavior clusters, 
we build a cluster-level action memory by averaging the visual-coupled action latents $\Psi$ produced by the world model. 
Concretely, we aggregate $\Psi_t^{(s)} \in \mathbb{R}^{L \times D}$ from all samples $s \in \mathcal{C}_m$ and compute 
per-cluster memory $\mathbf{M}_m$ as:
\begin{equation}
\mathbf{M}_m \;=\; \frac{1}{|\mathcal{C}_m|}\sum_{s \in \mathcal{C}_m} \Psi_t^{(s)} 
\;\in\; \mathbb{R}^{L \times D}.
\end{equation}
Then, stacking all cluster slices yields the full memory tensor $\mathbf{M}$ as
\begin{equation}
\mathbf{M} \;=\; \mathrm{Stack}\big(\mathbf{M}_1,\dots,\mathbf{M}_M\big) 
\;\in\; \mathbb{R}^{M \times L \times D}.
\label{equ:memory}
\end{equation}

\subsection{Stage3: E2E-AD Training in Heterogeneous Domains}
\label{sec:e2e_model}


In the second stage, we utilize the pre-trained world-model to yield two complementary representations: 
(i) the trajectory-guided visual prototypes $\{\mathbf{P}_m\}_{m=1}^{M}$, where $M$ denotes the number of behavior clusters, 
and (ii) the visual-coupled action memory $\mathbf{M} \in \mathbb{R}^{M \times L \times D}$.
In the third stage, the world model itself is not used during training or inference. Instead, we leverage the representations learned from the second stage to guide the E2E-AD model. Specifically, we use $\{\mathbf{P}_m\}$ to drive contrastive learning for trajectory-based representation alignment, while $\mathbf{M}$ retrieves action features to refine the predicted trajectory and ensure tight alignment between visual information and actions in a domain-agnostic manner, without relying on domain-specific information. 
The E2E-AD model is trained from scratch and follows a perception-free design~\cite{E2E_LAW_ICLR2025, zheng2025world4drive}, consisting of a visual encoder and a waypoint prediction module. By directly mapping raw sensor inputs to driving actions without explicit perception annotations, the model enables flexible learning across heterogeneous domains.


\noindent
\textbf{Contrastive Learning with Trajectory-guided Prototype.}
When training across heterogeneous domains, models often focus on inter-domain differences instead of planning-critical cues, which can yield suboptimal performance even with multi-dataset training. In contrast, the world model’s visual prototypes $\{\mathbf{P}_m\}$ align with trajectories rather than incidental domain factors, making them largely insensitive to domain bias. We therefore transfer this prototype knowledge to the E2E-AD model to steer learning toward trajectory-relevant information rather than domain-driven variations.

\begin{table*}[t]
\setlength{\tabcolsep}{2.6pt}
\centering
\renewcommand{\arraystretch}{1.1}
\caption{\textbf{Open-loop planning performance in nuScenes (S), NAVSIM (N), and Waymo (W) datasets.} The symbol $\dagger$ denotes an end-to-end autonomous driving model that uses perception annotations in addition to ground truth for planning.}
\vspace{-6pt}
\label{tab:open_loop_results}
\renewcommand\theadalign{c}
\renewcommand\cellset{\renewcommand\arraystretch{0.75}}
\resizebox{\linewidth}{!}{
\begin{tabular}{l|ccc|cccccccc|cccc|cccc|c}
\toprule
\multirow{3}{*}{Method} & \multicolumn{3}{c|}{\multirow{2}{*}{\thead{Train \\ Data}}} & \multicolumn{8}{c|}{nuScenes (S)} & \multicolumn{4}{c|}{NAVSIM (N)} & \multicolumn{4}{c|}{Waymo (W)} & 
\multicolumn{1}{c}{\multirow[c]{3}{*}{\makebox[-5pt][c]{\scriptsize\thead{Avg.\\of\\L2\\Avg.}}}}
\\ 
\cmidrule{5-20}
& & &
& \multicolumn{4}{c|}{L2 (m) $\downarrow$} & \multicolumn{4}{c|}{Collision (\%) $\downarrow$} & \multicolumn{4}{c|}{L2 (m) $\downarrow$} & \multicolumn{4}{c|}{L2 (m) $\downarrow$} & \\
\cmidrule{2-20}
 & S & N & W & 1s & 2s & 3s & Avg. & 1s & 2s & 3s & Avg. & 1s & 2s & 3s & Avg. & 1s & 2s & 3s & Avg. & \\
\hline
\hline
ST-P3$\dagger$~\cite{E2E_ST-P3}        
  & \cmark & - & - & 1.33 & 2.11 & 2.90 & \cellcolor{gray!20}2.11 
  & 0.23 & 0.62 & 1.27 & \cellcolor{gray!20}0.71 & 2.59 & 4.47
& 7.71 & \cellcolor{gray!20}4.92 & 3.76 & 6.27 & 9.72 & \cellcolor{gray!20}6.58 & \cellcolor{gray!20}4.54\\
UniAD$\dagger$~\cite{E2E_UniAD}         
  & \cmark & - & - & 0.48 & 0.74 & 1.07 & \cellcolor{gray!20}0.76 
  & 0.12 & 0.13 & 0.28 & \cellcolor{gray!20}0.17 & 2.04 & 5.44
& 12.80 & \cellcolor{gray!20}6.76 & 2.97 & 5.87 & 11.55 & \cellcolor{gray!20}6.80 & \cellcolor{gray!20}4.77\\
VAD-Tiny$\dagger$~\cite{E2E_VAD}  
  & \cmark & - & - & 0.20 & 0.38 & 0.65 & \cellcolor{gray!20}0.41 
  & 0.10 & 0.12 & 0.27 & \cellcolor{gray!20}0.16 & 5.02 & 8.43 & 11.87 & \cellcolor{gray!20}8.44 & 2.48 & 4.23 & 6.03 & \cellcolor{gray!20}4.25 & \cellcolor{gray!20}4.36\\
VAD-Base$\dagger$~\cite{E2E_VAD}   
  & \cmark & - & - & 0.17 & 0.34 & 0.60 & \cellcolor{gray!20}0.37 
  & 0.07 & 0.10 & 0.24 & \cellcolor{gray!20}0.14 & 4.25 & 7.15 & 10.09 & \cellcolor{gray!20}7.16 & 3.48 & 6.10 & 8.91 & \cellcolor{gray!20}6.16 & \cellcolor{gray!20}4.57\\
PARA-Drive$\dagger$~\cite{E2E_paradrive}
  & \cmark & - & - & 0.25 & 0.46 & 0.74 & \cellcolor{gray!20}0.48 
  & 0.14 & 0.23 & 0.39 & \cellcolor{gray!20}0.25 & - & - & - & \cellcolor{gray!20}- & - & - & - & \cellcolor{gray!20}- & \cellcolor{gray!20}- \\ 
SparseDrive$\dagger$~\cite{E2E_Sparsedrive} & \cmark & - & - & 0.29 & 0.58 & 0.96 & \cellcolor{gray!20}0.61 & 0.01 & 0.05 & 0.18 & \cellcolor{gray!20}0.08 & 1.81 & 2.98 & 4.17 &  \cellcolor{gray!20}2.99 & 3.48 & 5.62 & 7.64 & \cellcolor{gray!20}5.58 & \cellcolor{gray!20}2.88\\
DiffusionDrive$\dagger$~\cite{E2E_Diffusiondrive} & \cmark & - & -  & 0.27 & 0.54 & 0.90 & \cellcolor{gray!20}0.57 & 0.03 & 0.05 & 0.16 & \cellcolor{gray!20}0.08 & 1.95 & 3.15 & 4.30 & \cellcolor{gray!20}3.13 & 3.31 & 5.36 & 7.35 & \cellcolor{gray!20}5.34 & \cellcolor{gray!20}2.85 \\
\midrule
LTF~\cite{E2E_Tranfuser} 
  & \cmark & - & - & 0.45 & 0.66 & 0.99 & \cellcolor{gray!20}0.70 
  & 0.08 & 0.08 & 0.23 & \cellcolor{gray!20}0.13 & 0.49 & 1.04 & 1.82 & \cellcolor{gray!20}1.12 & 0.72 & 1.28 & 1.99 & \cellcolor{gray!20}1.33 & \cellcolor{gray!20}1.05\\
BEVPlaner~\cite{E2E_BEVPlanner}    
  & \cmark & - & - & 0.27 & 0.54 & 0.90 & \cellcolor{gray!20}0.57 
  & 0.04 & 0.35 & 1.80 & \cellcolor{gray!20}0.73 & 1.14 & 1.83 & 2.52 & \cellcolor{gray!20}1.83 & 2.25 & 3.76 & 5.29 & \cellcolor{gray!20}3.77 & \cellcolor{gray!20}2.06\\
BEVPlaner++~\cite{E2E_BEVPlanner}
 & \cmark & - & - & 0.16 & 0.32 & 0.57 & \cellcolor{gray!20}0.35 & 0.00 & 0.29 & 0.73 & \cellcolor{gray!20}0.34
  & 0.42 & 1.01 & 1.82 & \cellcolor{gray!20}1.08 & 0.47 & 1.00 & 1.67 &  \cellcolor{gray!20}1.05 & \cellcolor{gray!20}0.83\\
World4Drive~\cite{zheng2025world4drive} & \cmark & - & - &  0.23 & 0.47 &  0.81 & \cellcolor{gray!20}0.50 & 0.02 & 0.12 & 0.33 & \cellcolor{gray!20}0.16 & - & - & - & \cellcolor{gray!20}- & - & - & - & \cellcolor{gray!20}- & \cellcolor{gray!20}- \\
LAW~\cite{E2E_LAW_ICLR2025} & \cmark & - & - & 0.26 & 0.57 & 1.01 & \cellcolor{gray!20}0.61 & 0.14	& 0.21 & 0.54 & \cellcolor{gray!20}0.30 & 3.97 & 6.66 & 9.40 & \cellcolor{gray!20}6.68 & 4.02 & 6.67 & 9.31 & \cellcolor{gray!20}6.67 & \cellcolor{gray!20}4.65\\
\textbf{HEAT (Ours)} & \cmark & - & -  & 
   0.24 & 0.53 & 0.86 & \cellcolor{gray!20}0.54 &   0.08 & 0.16 & 0.44 & \cellcolor{gray!20}0.23 & 3.24 & 5.84 & 8.23 & \cellcolor{gray!20}5.77  & 3.74 &  6.02& 8.76 & \cellcolor{gray!20}6.17 & \cellcolor{gray!20}4.16 \\
\midrule
\midrule
LTF~\cite{E2E_Tranfuser}  & \cellcolor{het}\cmark & \cellcolor{het}\cmark & \cellcolor{het}\cmark & 1.05 & 2.89 & 4.95 & \cellcolor{gray!20}2.96 
  & 0.31 & 0.64 & 2.36 & \cellcolor{gray!20}1.10  & 0.80 & 1.99 & 3.40 & \cellcolor{gray!20}2.06 & 1.38 & 3.73 & 6.23 & \cellcolor{gray!20}3.78 & \cellcolor{gray!20}2.93\\

BEVPlaner~\cite{E2E_BEVPlanner} 
 & \cellcolor{het}\cmark & \cellcolor{het}\cmark & \cellcolor{het}\cmark & 1.71 & 2.87 & 4.06 & \cellcolor{gray!20}2.88 & 1.54 & 5.39 & 8.56 & \cellcolor{gray!20}5.16 & 2.10 & 3.55 & 5.05 & \cellcolor{gray!20}3.57 & 2.05 & 3.48 & 4.96 & \cellcolor{gray!20}3.50 & \cellcolor{gray!20}3.32\\
BEVPlaner++~\cite{E2E_BEVPlanner}
 & \cellcolor{het}\cmark & \cellcolor{het}\cmark & \cellcolor{het}\cmark &  0.17 & 0.34 & 0.62 & \cellcolor{gray!20}0.38 & 0.00 & 0.33 & 1.00 & \cellcolor{gray!20}0.44 & 0.10 & 0.32 & 0.71 & \cellcolor{gray!20}0.38 & 0.21 & 0.53 & 0.99 & \cellcolor{gray!20}0.58 & \cellcolor{gray!20}0.45\\
LAW~\cite{E2E_LAW_ICLR2025} & \cellcolor{het}\cmark & \cellcolor{het}\cmark & \cellcolor{het}\cmark & 0.43 & 0.79 & 1.26 & \cellcolor{gray!20}0.83 & 0.18 & 0.46 & 0.80 & \cellcolor{gray!20}0.48 & 0.74 & 1.20 & 1.69 & \cellcolor{gray!20}1.21 & 1.42 & 2.44 & 3.52 & \cellcolor{gray!20}2.46 & \cellcolor{gray!20}1.50 \\
  \textbf{HEAT (Ours)} & \cellcolor{het}\cmark & \cellcolor{het}\cmark & \cellcolor{het}\cmark & 
   0.15 & 0.32 & 0.58 & \cellcolor{gray!20}0.35 & 0.10 & 0.13 & 0.28 & \cellcolor{gray!20}0.17 &  0.07 & 0.20  & 0.40 & \cellcolor{gray!20}0.22 & 0.26 & 0.63 & 1.16 & \cellcolor{gray!20}0.68 & \cellcolor{gray!20}0.42 \\
\bottomrule
\end{tabular}}
\vspace{-16pt}
\end{table*}

To transfer knowledge from the constructed prototypes into the E2E-AD model, we adopt a contrastive learning objective. Given an input image $\mathcal{I}_t$, the E2E-AD encoder extracts visual tokens $\mathbf{V}_t=\{\mathbf{v}_t^1,\ldots,\mathbf{v}_t^L\}$ in a manner consistent with the world model. Since each image is assigned to a trajectory cluster $m$, we retrieve its corresponding prototype $\mathbf{P}_m$ as the positive and treat the remaining prototypes $\{\mathbf{P}_{m'} \mid m' \neq m\}$ as negatives.
The training objective enforces the extracted tokens to align closely 
with their positive prototype while remaining well separated from negative 
ones. Formally, for each token \(\mathbf{v}_t^l\), we define a contrastive 
loss with respect to the prototypes:  
\begin{equation}
\mathcal{L}_{\text{con}}^l = -\log 
\frac{\exp(\mathrm{sim}(\mathbf{v}_t^l, \mathbf{P}_m^l) / \tau)}
{\sum_{m'=1}^M \exp(\mathrm{sim}(\mathbf{v}_t^l, \mathbf{P}_{m'}^l) / \tau)},
\end{equation}
where \(\mathrm{sim}(\cdot,\cdot)\) denotes a cosine similarity function  and \(\tau\) is a temperature parameter. 
The overall contrastive objective is averaged across all tokens:
\begin{equation}
\mathcal{L}_{\text{con}} = \frac{1}{L} \sum_{l=1}^L \mathcal{L}_{\text{con}}^l.
\end{equation}
By aligning visual tokens with trajectory-guided prototypes, the E2E-AD model 
is encouraged to learn domain-invariant representations that capture 
behavioral intent rather than domain-specific appearance.

\noindent
\textbf{Episodic Memory-based Action Refinement.}
To further enhance decision consistency, we introduce an episodic memory module that utilizes the visual-coupled action memory $\mathbf{M}$ to refine action predictions. By retrieving action features from previously stored episodes that share similar visual contexts, the model can correct or adjust its current trajectory prediction, leading to more stable and domain-robust behavior.
Given surround-view visual tokens $\mathbf{V}_t=\{\mathbf{v}_{t}^{\,l}\}_{l=1}^{L}$, 
we first predict an initial future waypoint. 
A randomly initialized waypoint query $\mathbf{q}_{\mathrm{w}}^{init}$ attends over the spatial visual tokens $\mathbf{V}_t$:
\begin{equation}
\mathbf{q}^{\star}_{\mathrm{w}} \;=\; \mathrm{CrossAttn}\!\left(\mathbf{q}_{\mathrm{w}}^{init},\, \mathbf{V}_t\right),    
\end{equation}
and the initial waypoint is predicted by the head $f_{\mathrm{w}}^{init}$ as:
\begin{equation}
\hat{\mathbf{W}}_t \;=\; f_{\mathrm{w}}^{init}\!\left(\mathbf{q}^{\star}_{\mathrm{w}}\right),
\end{equation}
where
$\mathbf{\hat{W}}_t = \{\mathbf{\hat{w}}_t^1, \mathbf{\hat{w}}_t^2, \ldots, \mathbf{\hat{w}}_t^T\}$.
We then fuse the flattened initial waypoint with the spatial visual tokens by concatenation at token level,
forming an action-visual-fused features:
\begin{equation}
\mathbf{a}_t^l = \text{MLP}\big([\mathbf{v}_t^l, \mathbf{\hat{w}}'_t]\big), 
\quad l=1,\ldots,L,
\end{equation}
where $[\cdot,\cdot]$ denotes channel-wise concatenation and the resulting action-visual-fused feature $\mathbf{a}_t \in \mathbb{R}^{L \times D}$ is used for subsequent memory retrieval.

Given the visual-coupled action memory $\mathbf{M}\in\mathbb{R}^{M\times L\times D}$ and the action-visual-fused features from the E2E-AD model $\mathbf{a}_t\in\mathbb{R}^{L\times D}$, we project and $\ell_2$-normalize them to compute cosine attention between the features and the memory as:
\begin{equation}
\mathbf{Q} \;=\; \Phi(\mathbf{a}_t)\in\mathbb{R}^{L \times H},~ 
\mathbf{K} \;=\; \Phi(\mathbf{M})\in\mathbb{R}^{M\times L \times H},
\end{equation}
where $\Phi$ is the projection layer with normalization.
For memory attention, token-aligned similarities with a softmax over memory slots yield
\begin{equation}
\mathbf{A}_{l,m}\!=\!\operatorname{softmax}\!\big(\frac{\mathbf{Q}_{l}\!\cdot\!\mathbf{K}_{m,l}}{\tau}\big),~
\mathbf{C}_{l}\!=\!\sum_{m=1}^{M}\mathbf{A}_{l,m}\mathbf{M}_{m,l}.
\label{equ:attn}
\end{equation}
We then fuse the retrieved context with the action-fused features via a residual update as
$\mathbf{a}_{t}^\star = \mathbf{a}_{t} + \mathbf{C}$ and obtain the final waypoint prediction as:
\begin{equation}
\mathbf{W}_t  \;=\; f_{\mathrm{w}}(\mathrm{CrossAttn}\!\left(\mathbf{q}_{\mathrm{w}}^{init},\, \mathbf{a}_t^\star\right)). 
\end{equation}

\noindent
\textbf{Objective Function.}
Alongside the contrastive objective, the E2E-AD model is supervised by the 
ground-truth trajectories, \(\mathbf{\tilde{W}}_t\). The predicted waypoints $\hat{\mathbf{W}}_t, \mathbf{W}_t$ are 
regressed to match \(\mathbf{\tilde{W}}_t\) with an L1 loss. To facilitate training of perception-free E2E-AD framework~\cite{E2E_LAW_ICLR2025, zheng2025world4drive, E2E_WORLD_PLAN_chen2024drivinggpt}, a future latent reconstruction loss is adopted. The final loss combines all loss terms as:
\begin{equation}
\mathcal{L} = \mathcal{L}_{\text{traj}}^{final} +  \lambda_1 \, \mathcal{L}_{\text{traj}}^{init} + \lambda_2 \, \mathcal{L}_{\text{con}}+\lambda_3 \mathcal{L}_{recon}.
\label{equ:total_loss}
\end{equation}

\begin{table*}[t]
\centering
\caption{\textbf{Closed-loop planning performance in NAVSIM dataset.} NC: no at-fault collision. DAC: drivable area compliance. EP: ego progress. TTC: time-to-collision. Comf.: comfort. PDMS: the predictive driver model score.  C: Camera. L: LiDAR. In the training datasets column, S, N, and W denote nuScenes, NAVSIM, and Waymo, respectively. Models trained under the heterogeneous setting with all three datasets share the checkpoints used in Table~\ref{tab:open_loop_results}. $\dagger$ denotes a model that additionally uses perception annotations alongside planning ground truth.}
\label{tab:closed_loop}
\setlength\tabcolsep{9pt}
\renewcommand{\arraystretch}{0.7}
\vspace{-4pt}
\resizebox{1.0\linewidth}{!}{%
\begin{tabular}{l|c|ccc|ccccc|c}
\toprule
\multirow{2}{*}{Method} & \multirow{2}{*}{Input} & \multicolumn{3}{c|}{Train Data} 
& \multirow{2}{*}{NC $\uparrow$} & \multirow{2}{*}{DAC $\uparrow$} & \multirow{2}{*}{EP $\uparrow$} & \multirow{2}{*}{TTC $\uparrow$} & \multirow{2}{*}{Comf. $\uparrow$} & \multirow{2}{*}{PDMS $\uparrow$} \\
\cmidrule{3-5}
 &  & S & N & W   &  &  &  &  &  \\
\midrule
Human & - & - & - & - & 100 & 100 & 87.5 & 100 & 99.9 & \cellcolor{gray!20}94.8 \\
Constant Velocity & - & - & - & - & 69.9 & 58.8 & 49.3 & 49.3 & 100.0 & \cellcolor{gray!20}21.6 \\

\midrule

VADv2$\dagger$~\cite{E2E_vadv2} & C & - & \cmark & -  & 97.9 & 91.7 & 77.6 & 92.9 & 100.0 & \cellcolor{gray!20}83.0 \\
UniAD$\dagger$~\cite{E2E_UniAD}    & C & - & \cmark & -  & 97.8 & 91.9 & 78.8 & 92.9 & 100.0 & \cellcolor{gray!20}83.4 \\
PARA-Drive$\dagger$~\cite{E2E_paradrive} & C & - & \cmark & - & 97.9 & 92.4 & 79.3 & 93.0 & 99.8 & \cellcolor{gray!20}84.0 \\
Hydra-MDP$\dagger$~\cite{li2024hydra} & C \& L & - & \cmark & - & 98.3 & 96.0 & 78.7 & 94.6 & 100.0 & \cellcolor{gray!20}86.5 \\
WoTE$\dagger$~\cite{E2E_WORLD_PLAN_li2025end_wote} & C \& L & - & \cmark & - & 98.5 & 96.8 & 81.9 & 94.9 & 99.9 & \cellcolor{gray!20}88.3\\
\midrule

TransFuser~\cite{E2E_Tranfuser} & C \& L & - & \cmark & - & 97.7 & 92.8 & 79.2 & 92.8 & 100.0 & \cellcolor{gray!20}84.0 \\
LTF~\cite{E2E_Tranfuser}    & C & - & \cmark & - &   97.4 & 92.8 & 79.0 & 92.4 & 100.0 & \cellcolor{gray!20}83.8 \\


BEVPlanner++~\cite{E2E_BEVPlanner} & C &  - & \cmark & -  & 96.2 & 89.1 & 74.6 & 90.0 & 100.0 & \cellcolor{gray!20}79.0 \\
World4Drive~\cite{zheng2025world4drive}   & C & - & \cmark & -  & 97.4 & 94.3 & 79.9 & 92.8  & 100.0  & \cellcolor{gray!20}85.1 \\
LAW~\cite{E2E_LAW_ICLR2025}   & C & - & \cmark & - &  96.4 & 95.4 & 81.7 & 88.7 & 99.9 & \cellcolor{gray!20}84.6 \\
\midrule
\midrule


LTF~\cite{E2E_Tranfuser} & C & \cellcolor{het} \cmark & \cellcolor{het}\cmark & \cellcolor{het}\cmark & 92.4 & 83.8 & 36.4 & 87.6 & 81.3  &\cellcolor{gray!20}55.6 \\


BEVPlanner++~\cite{E2E_BEVPlanner} & C &  \cellcolor{het} \cmark & \cellcolor{het}\cmark & \cellcolor{het}\cmark  & 94.4 & 80.1 & 66.2 & 86.7 & 100.0 & \cellcolor{gray!20}69.6 \\

LAW~\cite{E2E_LAW_ICLR2025}   & C & \cellcolor{het} \cmark & \cellcolor{het}\cmark & \cellcolor{het}\cmark  & 93.7 & 83.9 & 69.7 & 86.4 & 98.4 & \cellcolor{gray!20}72.3 \\
\textbf{HEAT (Ours)} & C & \cellcolor{het} \cmark & \cellcolor{het}\cmark & \cellcolor{het}\cmark  & 97.7 &93.9 & 80.2 & 93.0 & 100.0 & \cellcolor{gray!20}85.2 \\
\bottomrule
\end{tabular}}
\vspace{-6pt}
\end{table*}


\section{Experiments}
\label{sec:experiment}

\subsection{Datasets}
To train and evaluate within the heterogeneous dataset setting, we employed three real-world datasets.

\noindent
\textbf{nuScenes~\cite{Benchmarks_NuScenes}:} The nuScenes dataset consists of 1,000 scenes. Following previous works~\cite{E2E_VAD, E2E_Diffusiondrive}, we evaluate planning using L2 displacement error, measuring the distance between predicted and ground-truth trajectories, and collision rate, quantifying trajectory intersections with agents.


\noindent
\textbf{NAVSIM~\cite{dauner2024navsim}:}
The NAVSIM benchmark provides a challenging setting for planning evaluation by resampling data to reduce trivial cases, thereby increasing scenario diversity and complexity. NAVSIM also supports closed-loop evaluation, capturing long-horizon behaviors and the cumulative effects of planning decisions. Performance is measured using the predictive driver model score (PDMS), which integrates no at-fault collisions (NC), drivable area compliance (DAC), time-to-collision (TTC), comfort, and ego progress (EP).

\noindent
\textbf{Waymo~\cite{waymo2025e2e}:} The Waymo end-to-end (E2E) driving dataset comprises 4,021 driving segments with eight camera views and ego-vehicle trajectories, with a particular emphasis on challenging long-tail scenarios. Unlike conventional autonomous driving datasets, it provides only camera inputs without additional sensor modalities and offers trajectory annotations alone as ground truth, without separate perception or prediction labels. We sampled one-tenth of the Waymo E2E dataset to prevent its large size relative to the other two datasets from dominating the overall data distribution.

\subsection{Implementation Details}
We employed Swin-Transformer-Tiny (Swin-T)~\cite{liu2021swin} as the backbone, with input images resized to 640$\times$384. Training uses a cosine annealing learning rate schedule~\cite{loshchilov2016sgdr} initialized at 5e-5. The model is optimized with AdamW~\cite{loshchilov2017decoupled} using a weight decay of 0.01, a batch size of 8, and is trained for 18 epochs on 8 RTX 3090 GPUs. The number of trajectory clusters $M$ is empirically set to 18. To enable learning and evaluation across heterogeneous datasets, we trained our model using eight camera views for NAVSIM and Waymo, while for nuScenes (which provides six views), we appended two blank images to match the same input dimension. Both waypoints and raw data were uniformly sampled at 2 Hz, and the model was designed to predict 4-second horizons, aligned with the NAVSIM closed-loop evaluation setting.

\subsection{Benchmark across Heterogeneous Domains}

To encourage multi-domain learning and fair comparison, we establish a two-part benchmark. Table~\ref{tab:open_loop_results} reports open-loop results. Models trained on the nuScenes benchmark are evaluated on nuScenes, NAVSIM, and Waymo. For NAVSIM and Waymo, we report only L2 displacement error. For methods without released checkpoints, we reproduce nuScenes results using the official code and evaluate those checkpoints on NAVSIM and Waymo. We also train these methods (LTF~\cite{E2E_Tranfuser}, BEVPlanner~\cite{E2E_BEVPlanner}, BEVPlanner++~\cite{E2E_BEVPlanner}, LAW~\cite{E2E_LAW_ICLR2025}) under a heterogeneous setting that jointly uses all three datasets and evaluate them in all domains. Because Waymo lacks perception labels, only perception-free variants are applicable. 
Table~\ref{tab:closed_loop} reports closed-loop results. For models trained under the heterogeneous setting, we use the same checkpoints as in Table~\ref{tab:open_loop_results} to assess whether E2E-AD models execute interaction-aware driving in closed-loop rollouts.


As shown in Table~\ref{tab:open_loop_results}, models perform strongly on the in-domain nuScenes dataset but suffer significant performance degradation under domain shifts, revealing substantial domain bias and highlighting the need for a unified model that can generalize across heterogeneous distributions.
In the heterogeneous training setup using all three datasets, models such as LTF fail to adapt effectively, even showing degraded performance compared to single-domain training. BEVPlanner++ and LAW exhibit substantial gains on NAVSIM and Waymo, but with trade-offs on nuScenes. Our proposed HEAT achieves the best overall open-loop performance, followed by BEVPlanner++.


From Table~\ref{tab:closed_loop}, we observe that BEVPlanner++ performs well in open-loop evaluation, but its closed-loop performance degrades significantly, suggesting a reliance on ego-status cues. 
LAW achieves a PDMS of 84.6 when trained on NAVSIM, but drops to 72.3 in the heterogeneous setting.
These results indicate that heterogeneous training improves generalization but may weaken domain-specific optimization due to feature separation. 
In contrast, our HEAT achieves strong performance in both open-loop and closed-loop evaluations, demonstrating robust learning and domain-invariant representations. 


\begin{table*}[t]
\centering
\caption{\textbf{Ablation study of the proposed method.} L2 and Collision Rates are reported as the average over 1s, 2s, and 3s. CLTP: Contrastive Learning with Trajectory-guided Prototype, EMAR: Episodic Memory-based Action Refinement.}
\label{tab:ablation}
\setlength\tabcolsep{3.2pt}
\renewcommand{\arraystretch}{0.9}
\resizebox{1.0\linewidth}{!}{%
\begin{tabular}{c|c|c|ccc|cccccc|cc|c}
\toprule
\multirow{2}{*}{ID} & \multicolumn{2}{c|}{Ablation Setting} & \multicolumn{3}{c|}{Train Data} 
& \multicolumn{6}{c|}{NAVSIM} & \multicolumn{2}{c|}{nuScenes} & \multicolumn{1}{c}{Waymo} \\
\cmidrule{2-15}
& CLTP & EMAR & S & N & W & NC $\uparrow$ & DAC $\uparrow$ & EP $\uparrow$ & TTC $\uparrow$ & Comf. $\uparrow$ & PDMS $\uparrow$ & L2 Avg. (m) $\downarrow$ & Col. Avg. (\%) $\downarrow$ &  L2 Avg. (m) $\downarrow$ \\
\midrule
1 &\xmark & \xmark & - & \cmark & - & 97.4 & 92.3 & 77.6 & 93.1 & 	100.0 & \cellcolor{gray!20}83.2 & \cellcolor{gray!20}5.98 & \cellcolor{gray!20}4.19 & \cellcolor{gray!20}6.43\\
\midrule
\midrule
2 &\xmark & \xmark  & \cellcolor{het} \cmark & \cellcolor{het}\cmark & \cellcolor{het}\cmark&
93.1 & 83.7 & 68.9 & 85.3 &	98.2 & \cellcolor{gray!20}71.4
 & \cellcolor{gray!20}0.90 & \cellcolor{gray!20}0.55 & \cellcolor{gray!20}2.48\\
3 &\cmark & \xmark  & \cellcolor{het} \cmark & \cellcolor{het}\cmark & \cellcolor{het}\cmark & 94.5 & 89.5 & 74.9 & 88.0 & 98.6 & \cellcolor{gray!20}78.0 & \cellcolor{gray!20}0.79 & \cellcolor{gray!20}0.44 & \cellcolor{gray!20}2.18
\\
4 & \xmark & \cmark  & \cellcolor{het} \cmark & \cellcolor{het}\cmark & \cellcolor{het}\cmark &  97.6 & 91.5 & 	77.7 & 93.0 & 100.0 & \cellcolor{gray!20}82.8 & \cellcolor{gray!20}0.42 & \cellcolor{gray!20}0.53 & \cellcolor{gray!20}0.70  \\
5 & \cmark & \cmark  & \cellcolor{het} \cmark & \cellcolor{het}\cmark & \cellcolor{het}\cmark &  
97.7 &93.9 & 80.2 & 93.0 & 100.0 & \cellcolor{gray!20}85.2 &\cellcolor{gray!20}0.35 & \cellcolor{gray!20}0.17 &\cellcolor{gray!20}0.68 \\
\bottomrule
\end{tabular}}
\vspace{-8pt}
\end{table*}

\begin{table}[t]
\centering
\caption{\textbf{Ablation study of the number of clusters.} The experiments were conducted under heterogeneous settings.}
\label{tab:num_cluster}
\vspace{-4pt}
\setlength\tabcolsep{2pt}
\begin{minipage}[t]{0.49\linewidth}
\centering
\resizebox{\linewidth}{!}{%
\begin{tabular}{c|c|cc|c}
\toprule
\multirow{2}{*}{$M$} & \multicolumn{1}{c|}{NAVSIM} & \multicolumn{2}{c|}{nuScenes} & \multicolumn{1}{c}{Waymo} \\
\cmidrule{2-5}
& PDMS $\uparrow$ & L2 Avg. $\downarrow$ & Col. Avg. $\downarrow$ & L2 Avg. $\downarrow$ \\
\midrule
1 & 72.3 & 0.74 & 0.52 & 1.40 \\
3 & 78.8 & 0.62 & 0.29 & 1.03 \\
6 & 84.8 & 0.38 & 0.17 & 0.72 \\
\bottomrule
\end{tabular}}
\end{minipage}
\hfill
\begin{minipage}[t]{0.48\linewidth}
\centering
\resizebox{\linewidth}{!}{%
\begin{tabular}{c|c|cc|c}
\toprule
\multirow{2}{*}{$M$} & \multicolumn{1}{c|}{NAVSIM} & \multicolumn{2}{c|}{nuScenes} & \multicolumn{1}{c}{Waymo} \\
\cmidrule{2-5}
& PDMS $\uparrow$ & L2 Avg. $\downarrow$ & Col. Avg. $\downarrow$ & L2 Avg. $\downarrow$ \\
\midrule
12 & 84.8 & 0.37 & 0.19 & 0.72 \\
18 & 85.2 & 0.35 & 0.17 & 0.68 \\
24 & 84.7 & 0.36 & 0.19 & 0.70 \\
\bottomrule
\end{tabular}}
\end{minipage}
\vspace{-20pt}
\end{table}


\subsection{Ablation Study and Discussion}

\noindent
\textbf{Effectiveness of the components.}
Table~\ref{tab:ablation} shows component ablation results. ID-1, trained only on NAVSIM without the proposed module, achieves strong NAVSIM closed-loop results but performs poorly on nuScenes and Waymo, highlighting the need for heterogeneous training. 
ID-2, trained heterogeneously yet still without the module, substantially improves on nuScenes and Waymo relative to ID-1, but this comes at the cost of NAVSIM performance, a trade-off we attribute to domain-specific bias.
In contrast, our approach improves performance across all domains, indicating successful trajectory-centric, domain-invariant learning. Notably, comparing ID-1 and ID-5, NAVSIM closed-loop performance also increases over the in-domain baseline, suggesting that heterogeneous training helps extract trajectory cues and enables the model to handle multi-domain data on a common representational footing.

\noindent
\textbf{Cluster number.} 
As shown in Table~\ref{tab:num_cluster}, the number of clusters $M$ influences the representation capacity of the model. Using a single cluster ($M=1$) leads to degraded performance, as a single latent space is insufficient to capture diverse driving behaviors. Increasing the number of clusters to $M=3$ provides improvements by roughly separating distinct motion patterns, such as straight driving, left turns, and right turns. As $M$ increases further, the performance becomes more stable, indicating that a richer set of trajectory-aware representations is beneficial. Although $M=18$ yields the best performance in our setting, the results remain consistent across a wide range of $M$, demonstrating that our method maintains robust performance under different cluster configurations.



\begin{figure}[t]
    \centering

    \begin{minipage}[t]{0.99\linewidth}
    \vspace{0pt} 
        \centering
        \includegraphics[width=\linewidth]{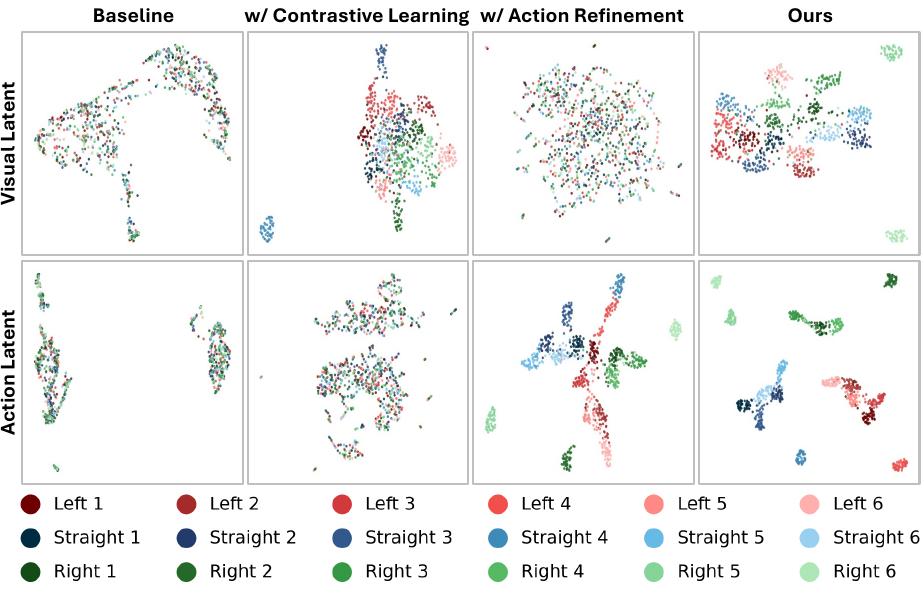}
        \vspace{-12pt}
        \captionof{figure}{UMAP~\cite{mcinnes2018umap} projections of the visual latent $\mathbf{V}_t$ and the action latent $\mathbf{a}_t$, visualized under different module configurations.}
        \label{fig:umap_viz}
    \end{minipage}
\end{figure}

\noindent
\textbf{Validation of Trajectory-Driven Learning.} 
As shown in Fig.~\ref{fig:umap_viz}, we visualize the feature spaces of the visual latent $\mathbf{V}_t$ and the action latent $\mathbf{a}_t$ while progressively adding modules to assess domain-invariant learning.
In the baseline without any modules, both visual and action latents cluster by data domain rather than by future trajectory, indicating a focus on domain-specific patterns. Introducing prototype-based contrastive learning yields trajectory-aligned structure in the visual latent. Refinement with the visual-coupled action memory then makes the action latent separable by trajectory as well. With both components, both visual and action latents form clear clusters by future trajectory.

\section{Conclusion}
\label{sec:conclusion}

In this work, we propose, to our knowledge for the first time, a method for training end-to-end autonomous driving (E2E-AD) across heterogeneous domains with differing data distributions, a capability that is essential for scalability and real-world deployment. To this end, we introduce a trajectory-driven learning framework based on prototypes and episodic memory, which delivers substantial gains in both open-loop and closed-loop evaluations under heterogeneous settings. We also present a new benchmark for heterogeneous E2E-AD, trained and evaluated across three datasets, to encourage further research.

\noindent
\textbf{Future work.}
While our work focuses on learning a unified model that performs consistently across multiple heterogeneous domains with diverse data distributions, achieving stable performance in this setting is itself a non-trivial challenge. Extending such capability to entirely unseen domains remains an important direction for future research, which we leave as future work.


\bibliographystyle{IEEEtran} 
\bibliography{main}          

@article{E2E_21_gaussian_NeurIPS2025_cho2025vr,
  title={VR-Drive: Viewpoint-Robust End-to-End Driving with Feed-Forward 3D Gaussian Splatting},
  author={Cho, Hoonhee and Kang, Jae-Young and Lee, Giwon and Yang, Hyemin and Park, Heejun and Jung, Seokwoo and Yoon, Kuk-Jin},
  journal={arXiv preprint arXiv:2510.23205},
  year={2025}
}

@inproceedings{note2eplan_8_CVPR2025_zhang2025carplanner,
  title={Carplanner: Consistent auto-regressive trajectory planning for large-scale reinforcement learning in autonomous driving},
  author={Zhang, Dongkun and Liang, Jiaming and Guo, Ke and Lu, Sha and Wang, Qi and Xiong, Rong and Miao, Zhenwei and Wang, Yue},
  booktitle={Proceedings of the Computer Vision and Pattern Recognition Conference},
  pages={17239--17248},
  year={2025}
}

@inproceedings{WORLD_DriveWM,
  title={Driving into the future: Multiview visual forecasting and planning with world model for autonomous driving},
  author={Wang, Yuqi and He, Jiawei and Fan, Lue and Li, Hongxin and Chen, Yuntao and Zhang, Zhaoxiang},
  booktitle={Proceedings of the IEEE/CVF Conference on Computer Vision and Pattern Recognition},
  pages={14749--14759},
  year={2024}
}

@article{mcinnes2018umap,
  title={Umap: Uniform manifold approximation and projection for dimension reduction},
  author={McInnes, Leland and Healy, John and Melville, James},
  journal={arXiv preprint arXiv:1802.03426},
  year={2018}
}

@article{E2E_WORLD_PLAN_chen2024drivinggpt,
  title={Drivinggpt: Unifying driving world modeling and planning with multi-modal autoregressive transformers},
  author={Chen, Yuntao and Wang, Yuqi and Zhang, Zhaoxiang},
  journal={arXiv preprint arXiv:2412.18607},
  year={2024}
}

@inproceedings{liu2021swin,
  title={Swin transformer: Hierarchical vision transformer using shifted windows},
  author={Liu, Ze and Lin, Yutong and Cao, Yue and Hu, Han and Wei, Yixuan and Zhang, Zheng and Lin, Stephen and Guo, Baining},
  booktitle={Proceedings of the IEEE/CVF international conference on computer vision},
  pages={10012--10022},
  year={2021}
}

@article{loshchilov2017decoupled,
  title={Decoupled weight decay regularization},
  author={Loshchilov, Ilya and Hutter, Frank},
  journal={arXiv preprint arXiv:1711.05101},
  year={2017}
}

@article{loshchilov2016sgdr,
  title={Sgdr: Stochastic gradient descent with warm restarts},
  author={Loshchilov, Ilya and Hutter, Frank},
  journal={arXiv preprint arXiv:1608.03983},
  year={2016}
}

@inproceedings{E2E_WORLD_PLAN_hassan2025gem,
  title={Gem: A generalizable ego-vision multimodal world model for fine-grained ego-motion, object dynamics, and scene composition control},
  author={Hassan, Mariam and Stapf, Sebastian and Rahimi, Ahmad and Rezende, Pedro and Haghighi, Yasaman and Br{\"u}ggemann, David and Katircioglu, Isinsu and Zhang, Lin and Chen, Xiaoran and Saha, Suman and others},
  booktitle={Proceedings of the Computer Vision and Pattern Recognition Conference},
  pages={22404--22415},
  year={2025}
}

@article{caesar2021nuplan,
  title={nuplan: A closed-loop ml-based planning benchmark for autonomous vehicles},
  author={Caesar, Holger and Kabzan, Juraj and Tan, Kok Seang and Fong, Whye Kit and Wolff, Eric and Lang, Alex and Fletcher, Luke and Beijbom, Oscar and Omari, Sammy},
  journal={arXiv preprint arXiv:2106.11810},
  year={2021}
}

@article{E2E_WORLD_PLAN_li2025end_wote,
  title={End-to-end driving with online trajectory evaluation via bev world model},
  author={Li, Yingyan and Wang, Yuqi and Liu, Yang and He, Jiawei and Fan, Lue and Zhang, Zhaoxiang},
  journal={arXiv preprint arXiv:2504.01941},
  year={2025}
}

@article{li2024hydra,
  title={Hydra-mdp: End-to-end multimodal planning with multi-target hydra-distillation},
  author={Li, Zhenxin and Li, Kailin and Wang, Shihao and Lan, Shiyi and Yu, Zhiding and Ji, Yishen and Li, Zhiqi and Zhu, Ziyue and Kautz, Jan and Wu, Zuxuan and others},
  journal={arXiv preprint arXiv:2406.06978},
  year={2024}
}

@inproceedings{WORLD_PLAN_ECCV2025_nemo,
  title={Neural volumetric world models for autonomous driving},
  author={Huang, Zanming and Zhang, Jimuyang and Ohn-Bar, Eshed},
  booktitle={European Conference on Computer Vision},
  pages={195--213},
  year={2024},
  organization={Springer}
}

@inproceedings{E2E_ST-P3,
  title={St-p3: End-to-end vision-based autonomous driving via spatial-temporal feature learning},
  author={Hu, Shengchao and Chen, Li and Wu, Penghao and Li, Hongyang and Yan, Junchi and Tao, Dacheng},
  booktitle={European Conference on Computer Vision},
  pages={533--549},
  year={2022},
  organization={Springer}
}

@inproceedings{E2E_Tranfuser,
  title={Multi-modal fusion transformer for end-to-end autonomous driving},
  author={Prakash, Aditya and Chitta, Kashyap and Geiger, Andreas},
  booktitle={Proceedings of the IEEE/CVF conference on computer vision and pattern recognition},
  pages={7077--7087},
  year={2021}
}

@article{E2E_TCP,
  title={Trajectory-guided control prediction for end-to-end autonomous driving: A simple yet strong baseline},
  author={Wu, Penghao and Jia, Xiaosong and Chen, Li and Yan, Junchi and Li, Hongyang and Qiao, Yu},
  journal={Advances in Neural Information Processing Systems},
  volume={35},
  pages={6119--6132},
  year={2022}
}

@inproceedings{E2E_Sparsedrive,
  title={Sparsedrive: End-to-end autonomous driving via sparse scene representation},
  author={Sun, Wenchao and Lin, Xuewu and Shi, Yining and Zhang, Chuang and Wu, Haoran and Zheng, Sifa},
  booktitle={2025 IEEE International Conference on Robotics and Automation (ICRA)},
  pages={8795--8801},
  year={2025},
  organization={IEEE}
}

@article{zheng2025world4drive,
  title={World4Drive: End-to-End Autonomous Driving via Intention-aware Physical Latent World Model},
  author={Zheng, Yupeng and Yang, Pengxuan and Xing, Zebin and Zhang, Qichao and Zheng, Yuhang and Gao, Yinfeng and Li, Pengfei and Zhang, Teng and Xia, Zhongpu and Jia, Peng and others},
  journal={arXiv preprint arXiv:2507.00603},
  year={2025}
}

@article{E2E_Diffusiondrive,
  title={DiffusionDrive: Truncated Diffusion Model for End-to-End Autonomous Driving},
  author={Liao, Bencheng and Chen, Shaoyu and Yin, Haoran and Jiang, Bo and Wang, Cheng and Yan, Sixu and Zhang, Xinbang and Li, Xiangyu and Zhang, Ying and Zhang, Qian and others},
  journal={arXiv preprint arXiv:2411.15139},
  year={2024}
}

@inproceedings{E2E_UniAD,
  title={Planning-oriented autonomous driving},
  author={Hu, Yihan and Yang, Jiazhi and Chen, Li and Li, Keyu and Sima, Chonghao and Zhu, Xizhou and Chai, Siqi and Du, Senyao and Lin, Tianwei and Wang, Wenhai and others},
  booktitle={Proceedings of the IEEE/CVF conference on computer vision and pattern recognition},
  pages={17853--17862},
  year={2023}
}

@inproceedings{E2E_Transfuser++,
  title={Hidden biases of end-to-end driving models},
  author={Jaeger, Bernhard and Chitta, Kashyap and Geiger, Andreas},
  booktitle={Proceedings of the IEEE/CVF International Conference on Computer Vision},
  pages={8240--8249},
  year={2023}
}

@inproceedings{E2E_VAD,
  title={Vad: Vectorized scene representation for efficient autonomous driving},
  author={Jiang, Bo and Chen, Shaoyu and Xu, Qing and Liao, Bencheng and Chen, Jiajie and Zhou, Helong and Zhang, Qian and Liu, Wenyu and Huang, Chang and Wang, Xinggang},
  booktitle={Proceedings of the IEEE/CVF International Conference on Computer Vision},
  pages={8340--8350},
  year={2023}
}

@inproceedings{E2E_BEVPlanner,
  title={Is ego status all you need for open-loop end-to-end autonomous driving?},
  author={Li, Zhiqi and Yu, Zhiding and Lan, Shiyi and Li, Jiahan and Kautz, Jan and Lu, Tong and Alvarez, Jose M},
  booktitle={Proceedings of the IEEE/CVF Conference on Computer Vision and Pattern Recognition},
  pages={14864--14873},
  year={2024}
}

@inproceedings{E2E_paradrive,
  title={Para-drive: Parallelized architecture for real-time autonomous driving},
  author={Weng, Xinshuo and Ivanovic, Boris and Wang, Yan and Wang, Yue and Pavone, Marco},
  booktitle={Proceedings of the IEEE/CVF Conference on Computer Vision and Pattern Recognition},
  pages={15449--15458},
  year={2024}
}

@article{dauner2024navsim,
  title={Navsim: Data-driven non-reactive autonomous vehicle simulation and benchmarking},
  author={Dauner, Daniel and Hallgarten, Marcel and Li, Tianyu and Weng, Xinshuo and Huang, Zhiyu and Yang, Zetong and Li, Hongyang and Gilitschenski, Igor and Ivanovic, Boris and Pavone, Marco and others},
  journal={Advances in Neural Information Processing Systems},
  volume={37},
  pages={28706--28719},
  year={2024}
}

@inproceedings{min2024driveworld,
  title={Driveworld: 4d pre-trained scene understanding via world models for autonomous driving},
  author={Min, Chen and Zhao, Dawei and Xiao, Liang and Zhao, Jian and Xu, Xinli and Zhu, Zheng and Jin, Lei and Li, Jianshu and Guo, Yulan and Xing, Junliang and others},
  booktitle={Proceedings of the IEEE/CVF conference on computer vision and pattern recognition},
  pages={15522--15533},
  year={2024}
}

@misc{waymo2025e2e,
  title        = {2025 Waymo Open Dataset Challenge: Vision-based End-to-End Driving},
  author       = {{Waymo Research}},
  year         = {2025},
  howpublished = {\url{https://waymo.com/open/challenges/2025/e2e-driving/}},
  note         = {Accessed: 2025-04-25}
}

@article{E2E_LAW_ICLR2025,
  title={Enhancing end-to-end autonomous driving with latent world model},
  author={Li, Yingyan and Fan, Lue and He, Jiawei and Wang, Yuqi and Chen, Yuntao and Zhang, Zhaoxiang and Tan, Tieniu},
  journal={arXiv preprint arXiv:2406.08481},
  year={2024}
}

@inproceedings{E2E_E2EAD_ICLR2025_navguided,
  title={Navigation-Guided Sparse Scene Representation for End-to-End Autonomous Driving},
  author={Li, Peidong and Cui, Dixiao},
  booktitle={The Thirteenth International Conference on Learning Representations}
}

@inproceedings{E2E_OccNet_ICCV2023,
  title={Scene as occupancy},
  author={Tong, Wenwen and Sima, Chonghao and Wang, Tai and Chen, Li and Wu, Silei and Deng, Hanming and Gu, Yi and Lu, Lewei and Luo, Ping and Lin, Dahua and others},
  booktitle={Proceedings of the IEEE/CVF International Conference on Computer Vision},
  pages={8406--8415},
  year={2023}
}

@article{E2E_ADMLP_archeive,
  title={Rethinking the open-loop evaluation of end-to-end autonomous driving in nuscenes},
  author={Zhai, Jiang-Tian and Feng, Ze and Du, Jinhao and Mao, Yongqiang and Liu, Jiang-Jiang and Tan, Zichang and Zhang, Yifu and Ye, Xiaoqing and Wang, Jingdong},
  journal={arXiv preprint arXiv:2305.10430},
  year={2023}
}

@INPROCEEDINGS{E2E_imitation_learning_ICRA2018,
  author={Codevilla, Felipe and Müller, Matthias and López, Antonio and Koltun, Vladlen and Dosovitskiy, Alexey},
  booktitle={2018 IEEE International Conference on Robotics and Automation (ICRA)}, 
  title={End-to-End Driving Via Conditional Imitation Learning}, 
  year={2018},
  volume={},
  number={},
  pages={4693-4700},
  doi={10.1109/ICRA.2018.8460487}}

@article{E2E_survey_IEEE2024,
  title={End-to-end autonomous driving: Challenges and frontiers},
  author={Chen, Li and Wu, Penghao and Chitta, Kashyap and Jaeger, Bernhard and Geiger, Andreas and Li, Hongyang},
  journal={IEEE Transactions on Pattern Analysis and Machine Intelligence},
  year={2024},
  publisher={IEEE}
}

@inproceedings{Benchmarks_NuScenes,
  title={nuscenes: A multimodal dataset for autonomous driving},
  author={Caesar, Holger and Bankiti, Varun and Lang, Alex H and Vora, Sourabh and Liong, Venice Erin and Xu, Qiang and Krishnan, Anush and Pan, Yu and Baldan, Giancarlo and Beijbom, Oscar},
  booktitle={Proceedings of the IEEE/CVF conference on computer vision and pattern recognition},
  pages={11621--11631},
  year={2020}
}

@article{vaswani2017attention,
  title={Attention is all you need},
  author={Vaswani, Ashish and Shazeer, Noam and Parmar, Niki and Uszkoreit, Jakob and Jones, Llion and Gomez, Aidan N and Kaiser, {\L}ukasz and Polosukhin, Illia},
  journal={Advances in neural information processing systems},
  volume={30},
  year={2017}
}

@inproceedings{E2E_vadv2,
  title={VADv2: End-to-End Autonomous Driving via Probabilistic Planning},
  author={Jiang, Bo and Chen, Shaoyu and Gao, Hao and Liao, Bencheng and Zhang, Qian and Liu, Wenyu and Wang, Xinggang},
  booktitle={The Fourteenth International Conference on Learning Representations},
  year={2026}
}

\vfill

\end{document}